\documentclass[preprint,12pt,authoryear]{elsarticle}
\usepackage{xcolor}

\usepackage{amssymb}
\usepackage{amsmath}
\usepackage{makecell}

\journal{International Journal of Medical Informatics}

\begin{document}

\begin{frontmatter}


\title{Precision Radiotherapy via Information Integration of Expert Human Knowledge and AI Recommendation to Optimize Clinical Decision Making}

\author[label1]{Wenbo Sun}
\ead{sunwbgt@umich.edu}
\author[label2]{Dipesh Niraula}
\ead{Dipesh.Niraula@moffitt.org}
\author[label2]{Issam El Naqa}
\ead{Issam.ElNaqa@moffitt.org}
\author[label3]{Randall K Ten Haken}
\ead{rth@med.umich.edu}
\author[label4]{Ivo D Dinov}
\ead{dinov@umich.edu}
\author[label3]{Kyle Cuneo}
\ead{kcuneo@umich.edu}
\author[label1]{Judy (Jionghua) Jin}
\ead{jhjin@umich.edu}
\address[label1]{Department of Industrial and Operations Engineering, University of Michigan, Ann Arbor}
\address[label2]{Department of Machine Learning, H. Lee Moffitt Cancer Center \& Research Institute, Tampa, FL}
\address[label3]{Department of Radiation Oncology, University of Michigan, Ann Arbor}
\address[label4]{Department of Computational Medicine and Bioinformatics, University of Michigan, Ann Arbor}

\begin{abstract}
In the precision medicine era, there is a growing need for precision radiotherapy where the planned radiation dose needs to be optimally determined by considering a myriad of patient-specific information in order to ensure treatment efficacy. Existing artificial-intelligence (AI) methods can recommend radiation dose prescriptions within the scope of this available information. However, treating physicians may not fully entrust the AI's recommended prescriptions due to known limitations or when the AI recommendation may go beyond physicians' current knowledge. This paper lays out a systematic method to integrate expert human knowledge with AI recommendations for optimizing clinical decision making. Towards this goal, Gaussian process (GP) models are integrated with deep neural networks (DNNs) to quantify the uncertainty of the treatment outcomes given by physicians and AI recommendations, respectively, which are further used as a guideline to educate clinical physicians and improve AI models performance. The proposed method is demonstrated in a comprehensive dataset where patient-specific information and treatment outcomes are prospectively collected during radiotherapy of $67$ non-small cell lung cancer patients and retrospectively analyzed.
\end{abstract}


\begin{highlights}
\item Develop an integrative system to help physicians design radiotherapy by combining human knowledge and AI recommendation.
\item Quantify the uncertainty of the treatment outcome based on black-box AI algorithms and physicians' prescriptions.
\item Use the data analytic results to educate physicians and improve the AI recommendations.
\item Demonstrate the proposed method using real patient data from radiotherapy.
\end{highlights}



\end{frontmatter}


\section{Introduction}\label{s:intro}
There is a growing need for precision radiotherapy since the precision medicine era was highlighted by the Joint American Society for Radiation Oncology (ASTRO) / National Cancer Institute (NCI) workshops in Bethesda, MD \citep{benedict2016overview, ashton2018dual}. Since the treatment outcome of a radiotherapy depends on the patient information and dose prescription, the planned radiation dose prescription should be optimized based on the patient-specific information to achieve the optimal treatment outcomes \citep{el2018prospects} for precision radiotherapy. The radiotherapy effectiveness is often evaluated by checking whether the local control (LC) of the tumor is assured and the main side effects such as radiation-induced pneumonitis (RP2) is mitigated when the patients complete all planned radiotherapy treatments. It is desired that the tumor tissue is LC'ed while the normal tissue gets least disturbed due to RP2. As the treatment outcome has its inherent randomness, it is expected to maximize the probability of LC and minimize the probability of RP2.

\begin{figure}[!htbp]
    \centering
    \includegraphics[height=1.7in]{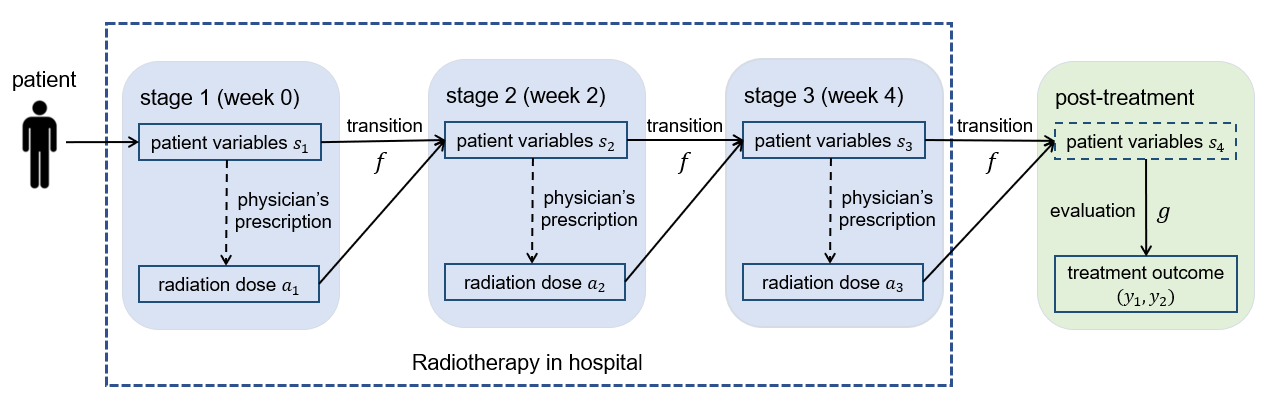}
    \caption{Illustration of the dose escalation protocol utilized in the study}
    \label{f:patient}
\end{figure}

The radiotherapy plan is designed following 3 successful dose escalation protocols \citep{luo2018multiobjective} as shown in Figure~\ref{f:patient}. Each radiotherapy plan is divided into $3$ stages, in which the patient's state variables (simplified as ``patient variables'' in the later context) $s_1$, $s_2$ and $s_3$ are consistently monitored. In each stage, fractions of radiation doses with dose per fraction $a_1$, $a_2$, and $a_3$ were delivered at the three stages, respectively. In the protocols, physicians fix $a_1$ and $a_2$ as $2$ Gy/frac in the first two stages, and adjust $a_3$ based on the patient variables $s_3$. After the three-stage radiotherapy plan, the treatment outcomes are evaluated by two binary variables $y_1\in\{0,1\}$ and $y_2\in\{0,1\}$, where $y_1=1$ indicates LC while $y_2=1$ indicates RP2. In practice, $a_3$ is often determined based on physicians' expert human knowledge and may not always achieve the optimal treatment outcome. Such a limitation motivates the development of the precision radiotherapy techniques.

The objective of the precision radiotherapy process is to adjust $a_3$ for optimal treatment outcome, that is to optimize the trade-off between the high probability of LC (denoted by $Prob[y_1=1]$) and low probability of RP2 (denoted by $Prob[y_2=1]$). The optimally planned radiation dose prescription $a_3$ (simplified called prescription in latter context) is usually searched via artificial intelligence (AI) approaches. There are two main types of approaches: direct methods and indirect methods. The direct methods have been proposed by \citet{zhao2012estimating}, which search for the optimal prescription directly based on the weighted outcomes in learning (called O-learning) without requiring a prediction model for the potential outcomes of future prescriptions. Thus, there is no need to fit a prediction model, and the process is model free. In contrast, indirect methods take a two-step analysis, i.e., first to fit a model that can accurately predict the outcomes in the prescription space, and then to search for an optimal prescription based on the predicted outcomes, that is a model based approach to improve performance. Q-learning is one of the most popularly used also in these indirect, or model-based methods \citep{qian2011performance, chakraborty2013statistical, moodie2014q}. Considering the patient variables are dynamically changing over different stages of radiotherapy, it is more appropriate to use the indirect method for precision radiotherapy, that is, to predict the probability of the treatment outcome $\left(Prob[y_1=1],Prob[y_2=1]\right)$, based on which optimal prescription $a_3$ is obtained by optimizing the predicted probabilities $\left(Prob[y_1=1],Prob[y2=1]\right)$.

To develop the indirect methods, the treatment outcome will be predicted via existing machine learning approaches such as penalized least square \citep{qian2011performance}, Bayesian networks \citep{pearl2009causal}, deep neural networks \citep{goodfellow2016deep}, etc. Since the effect of radiotherapy among multiple stages is a complex nonlinear function on the patient variables, deep neural networks are usually selected \citep{tseng2017deep}. When the Q-learning is integrated with the deep neural network, it is also known as deep Q network. A major limitation of deep Q-learning is its instability due to limited training data – a small error in prediction of the treatment outcome may significantly affect the final AI recommendation, especially when the sample size of patients is very limited as typically the case. Another concern of the deep Q-learning is the interpretability – the deep neural network is usually constructed as a black box, making it difficult to interpret the relationship between the predictors and outcomes in a straightforward manner. 

\begin{figure}[!htbp]
    \centering
    \includegraphics[height=1.9in]{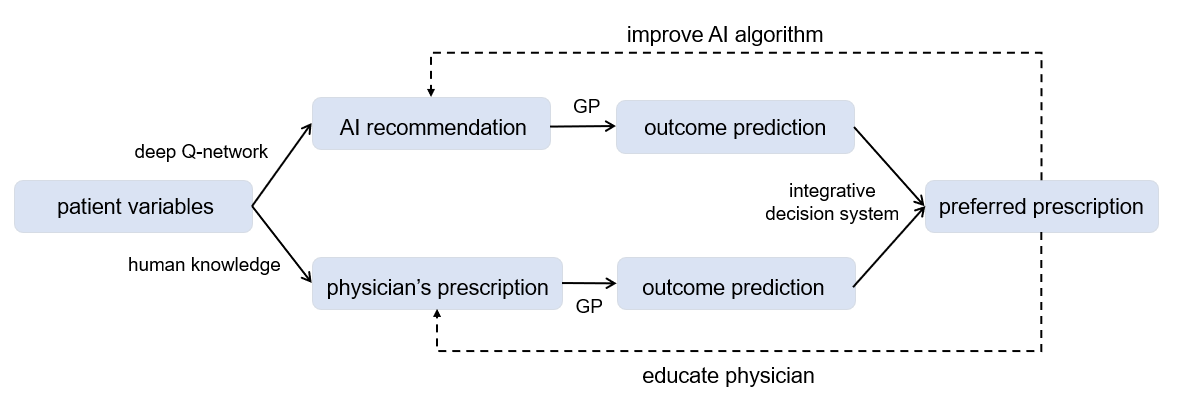}
    \caption{Information flow and decision support system framework}
    \label{f:flowchart}
\end{figure}

Ideally, when the Q-learning approach always provides a prescription that is close to the oracle’s benchmark, clinical physicians feel comfortable to follow the AI recommendation. However, when the treatment outcome is not accurately predicted due to small and incomplete datasets, learning from the results of practitioners' decisions or biased inference and evaluation processes \citep{rich2019lessons}, physician's prescription may result in preferable treatment outcomes than those are aroused by following the AI recommendation. In this paper, instead of seeking the origin of the prediction bias, we aim to develop a systematic approach to guide the clinical physicians when their prescriptions are different from the AI recommendations. We propose to use the commonly used tool for uncertainty quantification, Gaussian process (GP) model \citep{rasmussen2003gaussian}, to compensate the gap between the AI predictions of the treatment outcomes and their true values under the computer model calibration framework \citep{kennedy2001bayesian}. An integrative decision system will be developed: when the physicians' prescriptions are desired, the system suggests how the AI algorithm can be improved. When the AI recommendations provide better treatment outcomes, the system helps physicians make better decision for future patients. The flowchart of the proposed method is provided in Figure~\ref{f:flowchart}.

\section{Notations and problem formulation}\label{s:notations}
To start with, we would like to introduce some mathematical notations that will be used to elaborate the proposed method. Let $T$ denote the total number of stages of radiotherapy ($T=3$ in the dose escalation protocols \citep{luo2018multiobjective}). Based on Figure~\ref{f:patient}, for a specific patient, the patient variables $s_t$ are observed at stage $t=1,...,T$ when the patient is in the hospital. When the patient exits from the hospital, the treatment outcome $(y_1,y_2)$, instead of the patient variables $s_{T+1}$, is recorded. At stages $t=1,...,T-1$, radiation treatments with a known dose per fraction are delivered to the patient. At stage $T$, the dose per fraction $a_{T}$ needs to be adjusted based on $s_{T}$ to achieve the optimal treatment outcome. 

As was mentioned in Section~\ref{s:intro}, the optimal treatment outcome refers to the trade-off between the high LC probability and low RP2 probability. Here we modify the ``P+'' reward function in \cite{luo2018multiobjective} to the smoothed version $R$, which is defined as:
\begin{equation}
    R=-10\times\left(\left(1-Prob\left[y_1=1\right]\right)^8+\left(\frac{Prob\left[y_2=1\right]}{0.57}\right)^8\right)^{1/8}+3.281.
\label{eq:contour}
\end{equation}
Given the reward function, we define the action value function $Q$ as a Markov Decision Process (MDP) as:
\begin{equation}
    Q_\pi\left(s_{T},a_{T}\right)=\mathbb{E}_\pi\left[R_{T+1}+\gamma R_{T+2}+\gamma^2 R_{T+3}+...\middle|s_T,a_T\right],
\end{equation}
where $R_{t}$ represents the reward at time $t$ and $\gamma$ denotes the discount factor. By setting $R_{T+1}=R$, we can further derive the action value function as:
\begin{multline}
    Q_\pi\left(s_{T},a_{T}\right)=\mathbb{E}\left[R\middle|s_T,a_T\right] \\ +\gamma \int_{s_{T+1}} d s_{T+1} p\left[s_{T+1}\middle|s_T,a_T\right] \int_{a_{T+1}}da_{T+1}\pi\left[a_{T+1}\middle| s_{T+1}\right]Q_\pi\left(s_{T+1},a_{T+1}\right),
    \label{eq:opt}
\end{multline}
with $p$ representing the probability density function, policy $\pi\left[a\middle|s\right]$ representing probability density of choosing $a$, and $s_{T+1}$ denoting the true, yet unknown patient variables after treatment. With $Q$ defined, the optimization problem is formulated as:
\begin{equation}
	a_{T}^*=\pi^*\left(s_{T}\right)=arg\,max_{a_{T}\in\mathcal{A}} Q\left(s_{T},a_{T}\right).
\end{equation}
Solving the above optimization problem yields the optimal dose decision for AI recommendation.

Equation~(\ref{eq:opt}) involves two unknown terms that should be modeled and estimated, $p\left[s_{T+1}\middle| s_{T},a_{T}\right]$ and $E\left[R\middle| s_{T+1}\right]$. The former term is named by ``transition function'' $f$, which updates the transition of the patient variables from the previous stage to the current stage under given dose per fraction $a_{T}$. The latter term evaluates the reward function based on $Prob\left[y_1=1\right]$ and $Prob\left[y_2=1\right]$, which further requires an ``evaluation function'' $g$ that predicts the probability of the treatment outcome based on $s_T$. The relationship between the two functions $f$ and $g$ and the other variables is illustrated in Figure~\ref{f:patient}. Quantifying $f$ and $g$ requires a clinical trial dataset that collects radiotherapy records. In the dataset, the patient variables are stored in $S=\left\{s_{i,t};i=1,...,n;t=1,...,T\right\}$ where $s_{i,t}\in \mathbb{R}^q$ represent the patient $i$’s patient variables at stage $t$ with dimension of $q$. At each stage $t\in\{1,...,T\}$, the physicians deliver radiation treatments with dose per fraction $a_{i,t}$ to the patient $i$. When the radiotherapy reaches to the end, the treatment outcome are recorded as $y_{i,1}$ and $y_{i,2}$. Suppose the estimates are denoted by $\hat{f}$ and $\hat{g}$. The reward function $R$ in Equation~\ref{eq:opt} can be estimated as $\hat{g}\left(\ \hat{f}\left(s_{T},a_{T}\right)\right)$. The AI recommendation is hereby computed as
\begin{equation}
{\hat{a}}_{T}=arg\,max_{a_{T}\in\mathcal{A}}\ \hat{g}\left(\ \hat{f}\left(s_{T},a_{T}\right)\right).
\label{eq:opt_est}
\end{equation}
The risks of following the AI recommendations in the real practice are caused by inaccurate $\hat{f}$ and $\hat{g}$ – when these two estimators do not well approximate the true $f$ and $g$, the resultant ${\hat{a}}_{T}$ will deviate from the true optimal prescription $a_{T}^*$, and may lead to poor treatment outcome. To help physicians determine the prescription with a given AI recommendation, a methodology will be developed in the following framework: firstly, the uncertainty of the treatment outcome is quantified through the GP modelling of $f$ and $g$. Secondly, a comparison between the treatment outcomes is conducted via a hypothesis testing, assisting the physician to determine whether to use their own prescription or trust in the AI recommendation. Lastly, the comparison results will be fed back to improve both the human expert knowledge and AI recommendation.

\section{Methodology}\label{s:method}
\subsection{Estimation of the transition function}\label{ss:transition}
The transition function $f$ provides the one-step-ahead prediction of the patient variables given the current patient variables and the corresponding prescription. $\left\{\left(s_t,a_t\right),t=1,...,T-1\right\}$ are treated as predictors and $\left\{s_{t+1},t=1,... T-1\right\}$ are considered as the corresponding responses. Once $f$ is trained, the state variables at the final stage, $s_{T+1}$, can be predicted based on $s_{T}$ and $a_{T}$.  
A point estimator of $f$ can be trained as a DNN by utilizing the data generated from a generative adversarial network (GAN) as in \cite{goodfellow2014generative}. However, since DNN is a black box, there is no straightforward way to quantify the uncertainty of $f$, which is crucial in evaluating the accuracy of the resultant AI recommendation. To do this, we attach a GP-based bias term to the DNN following the computer model calibration framework \cite{kennedy2001bayesian} as:
\begin{equation}
    s_{t+1}=f\left(s_t\right)=\eta_s\left(s_t\right)+\delta_s\left(s_t\right),t=1,...,T,
\end{equation}
where $\eta_s\left(s_t\right)$ represents the point estimator obtained from DNN, and $\delta_s\left(s_t\right)$ represents the model bias. The bias $\delta_s\left(s_t\right)=\{\delta_s^{\left(1\right)}\left(s_t\right),...,\delta_s^{\left(q\right)}\left(s_t\right)\}$ is a $q$-dimensional random function, each dimension of which is assumed to follow a GP as:
\begin{equation}
    \delta_s^{\left(j\right)}\left(\cdot\right)\sim\mathcal{GP}\left(0,\frac{1}{\lambda_\delta^{\left(j\right)}}c_\delta^{\left(j\right)}\left(\cdot,\cdot\right)\right),j=1,...,q,
\end{equation}
where $c_\delta^{\left(j\right)}\left(\cdot,\cdot\right)$ is the covariance function and $\lambda_\delta^{\left(j\right)}$ is the precision parameter of GP. The covariance function is assumed to take the squared exponential (SE) form as:
\begin{eqnarray}
    c_\delta^{\left(j\right)}\left(\left(s,a\right),\left(s^\prime,a^\prime\right)\right)&=&\prod_{k=1}^{q}\exp{\left(-\beta_{\delta,k,j}\left(s^{\left(k\right)}-s^{\left(k\right)\prime}\right)^2\right)}\nonumber\\
    &\times& \exp{\left(-\beta_{\delta,q+1,j}\left(a-a^\prime\right)^2\right)}.
    \label{eq:se_transition}
\end{eqnarray}
Here $s^{(k)}$ denotes the $k-$th dimension of the vector $s$. Following the parameter estimation and Gaussian process prediction techniques shown in \ref{aptransition}, we are able to generate prediction of $s_{T+1}$, denoted by $\widehat{S}_{T+1}$. The prediction $\widehat{S}_{T+1}$ can be further used to predict the reward function.

\subsection{Estimation of the evaluation function}
The evaluation function $g$ predicts the reward function $R$ based on the predictions of $s_{T+1}$. Since $R$ is determined by $Prob\left[y_1=1\right]$ and $Prob\left[y_2=1\right]$, it is sufficient to predict the two probabilities $Prob\left[y_1=1\right]$ and $Prob\left[y_2=1\right]$, or equivalently, the logit function of the two probabilities, denoted by $h_1$ and $h_2$, respectively. Here we assume the two random functions $h_1$ and $h_2$ to follow GPs as
\begin{eqnarray}
h_1&\sim&\mathcal{GP}\left(0,\frac{1}{\lambda_{h_1}}c_{h_1}\left(\cdot,\cdot\right)\right),\nonumber\\
h_2&\sim&\mathcal{GP}\left(0,\frac{1}{\lambda_{h_2}}c_{h_1}\left(\cdot,\cdot\right)\right),
\end{eqnarray}
where $\lambda_{h_1}$ and $\lambda_{h_2}$ are the precision parameters, $c_{h_1}\left(\cdot,\cdot\right)$ and $c_{h_1}\left(\cdot,\cdot\right)$ are the covariance functions constructed under the squared exponential (SE) form: 
\begin{equation}
c_{h_j}\left(s,s\prime\right)=\prod_{k=1}^{q}exp{\left(-\beta_{h,k,j}\left(s^{\left(k\right)}-s^{\left(k\right)^\prime}\right)^2\right)}.    
\end{equation}

However, the two GPs cannot be directly trained because the logit of $LC$ and $RP2$ in the dataset, denoted by $H_1$ and $H_2$, are not observed. Instead, we only observe the binary labels $Y_1=\{y_{i,1};i=1,...,n\}$ and $Y_2=\{y_{i,2};i=1,...,n\}$. As a solution, we utilize the Laplace Approximation technique \citep{tierney1986accurate} to provide the estimation of GP parameters and the prediction of the $LC$ and $RP2$ probabilities. Detailed derivations can be found in \ref{apevaluation}. Combination of the evaluation function and transition function generate the prediction of $R$ for any given patient variable $s_{T}$ and dose prescription $a_{T}$, which can be used to search for the optimal dose prescription for AI recommendation via the deep Q-learning algorithm.

\subsection{Integration of physician's prescription and AI recommendation}\label{ss:integrate}
Now we aim to develop an integrative system that can help make prescription decisions and improve both physician's prescription and AI recommendation in the future. The prerequisite is to quantify the uncertainty of $R$ for given $s_{T}$ and $a_{T}$, that is to develop a confidence interval estimator of the composite function $g\left(f\left(\cdot\right)\right)$. Detailed derivations can be found in \ref{apcombination}.

Next, we generate the predictive distribution of the reward function $R$ via Monte Carlo simulations since the reward function is not linear with respect to the LC and RP2 probabilities. We firstly generate simulated $Prob[LC]$ and $Prob[RP2]$ from Equation~(\ref{eq:sim}). Then compute the simulated reward function based on its definition in Equation~(\ref{eq:contour}). Suppose the AI recommendation is $a^A(s_{T-1})$ for patient variable $s_{T}$, the physician's prescription is $a^P(s_{T})$. We denote the simulated reward functions based on the AI recommendation by
\begin{equation}
    \tilde{r}\left(s_{T},a^A\left(s_{T}\right)\right).
\end{equation}
The simulated reward functions based on the physician's prescription is denoted by
\begin{equation}
    \tilde{r}\left(s_{T},a^P\left(s_{T}\right)\right).
\end{equation} 
Given the Monte Carlo simulated reward functions, a student-t test is conducted to test whether 
\begin{equation}
    \tilde{r}\left(s_{T},a^A\left(s_{T}\right)\right)
\end{equation} 
are significantly greater than
\begin{equation}
    \tilde{r}\left(s_{T},a^P\left(s_{T}\right)\right).
\end{equation}
The system selects the AI recommendation when the null hypothesis is rejected, or equivalently, the p-value is smaller than $0.05$. Otherwise, the system will recommend physicians to follow their original prescription.

Thirdly, we would like to utilize the hypothesis testing results to improve the physician's prescription and AI recommendation for future patients. For physicians, they need to know the amount of dose that need to be compensated comparing to that from their human expert knowledge. To do this, we assume the physician's dose bias $\delta_a\left(s_{T}\right)=a^A\left(s_{T}\right)-a^P\left(s_{T}\right)$ to follow a GP with considering $s_{T}$ as inputs. The GP can be trained following the same procedure as in subsection~\ref{ss:transition}. Here we only utilize the samples whose corresponding p-values are smaller than $0.05$, implying that the AI recommendation provides better treatment outcome than the original physician's prescription at those samples. 

When AI recommendations are not reliable, physicians should be warned. To do this, the confidence intervals of $Prob[y_j=1|s_T,a^A(s_T)]$ and $Prob[y_j=2|s_T,a^A(s_T)]$ are visualized for physicians. The large confidence intervals imply that the AI recommendations are not reliable. In these cases, it is recommended that physicians should insist on their own prescription and be careful when delivering the dose prescription. The detailed guidelines will be provided in Section~\ref{s:results}.

\section{Case study in radiotherapy of lung cancer}\label{s:results}
In this section, the proposed method is demonstrated in a dataset collecting the radiotherapy records of $67$ patients in $4$ weeks. The dataset is part of the clinical trial data with real patients undergoing treatment. Part of the original dataset was published here in \citet{kong2017effect}. In the dataset, a total of $250$ potential patient variables are collected as the patient information before and during radiotherapy. They include dosimetric variables, clinical factors, circulating microRNAs, single-nucleotide polymorphisms (SNPs), circulating cytokines, positron emission tomography (PET) imaging radiomics features, etc. To reduce the dimension of patient variables to a feasible number, Bayesian Network technique \citep{luo2018multiobjective, luo2017unraveling} is applied to search for the most related patient variables to LC and RP2 where the selected variables are identified as the Markov blankets (MBs) of LC and RP2. As a result, 12 out of 250 patient variables are selected as the predictors for LC and RP2. They are il4, il10, il5, ip10, MTV, GLSZM\_LZLGE, GLSZM\_ZSV, Tumor\_gEUD, Lung\_gEUD, Rs2234671, Rs238406, Rs1047768. To avoid sparsity in the space of training dataset, extreme values are removed by truncating all the patient variables at their $70\%$ quantiles. Then the patient variables are scaled into $[0, 1]$ to train the transition and evaluation functions. Detailed descriptions of the $12$ selected patient variables can be found in \ref{appB}.

After pre-processing the data, we start to train the transition function $f$ as a combination of DNN and GP. When training the DNN, GAN is implemented to augment the sample size to ensure the fitting accuracy. A comparison between the predictions from the original deep neural network and the Gaussian process model is shown in Table~\ref{t:transition}. The last three patient variables are not predicted because they are constant. The cross-validation mean square errors (MSE) of the week 2 and week 4 data are calculated as the evaluation criterion. The relative improvement is computed as
\begin{equation}
    RI_j=\frac{\sum_{t=1}^2\|\hat\mu_{t+1}^{(j)}(S_t,A_t)-\eta_s^{(j)}(S_t,A_t)\|^2}{\sum_{t=1}^2\|\eta_s^{(j)}(S_t,A_t)-S_{t+1}^{(j)}\|^2}.
\end{equation}
Since DNN already achieves a relatively high accuracy, the GP model does not significantly improve the accuracy. However, the Gaussian process model provides the predicted confidence interval of the patient variables, which can be further used to quantify the uncertainty of the treatment outcome in the next steps.
\begin{table*}
\centering
\caption{Cross-validation MSE of the transition function}
\label{t:transition}
\begin{tabular}{c|c c c}
\hline
Response & DNN & GP & Relative Improvement\\
\hline
il4 & 0.000287 & 0.000110 & $72.29\%$\\
il10 & 0.000750 & 0.000691 & $64.16\%$\\
il5 & 0.000394 & 0.000367 & $28.02\%$\\
ip10 & 0.000038 & 0.000002 & $89.79\%$\\
mtv & 0.010464 & 0.009343 & $44.60\%$\\
GLSZM\_LZLGE & 0.014043 & 0.016114 & $31.86\%$\\
GLSZM\_ZSV & 0.001932 & 0.002155 & $21.31\%$\\
Lung gEUD & 0.000182 & 0.000100 & $58.10\%$\\
Tumor gEUD & 0.000480 & 0.000442 & $26.39\%$\\
\hline
\end{tabular}
\end{table*}

\begin{figure}
    \centering
    \includegraphics[height=2.7in]{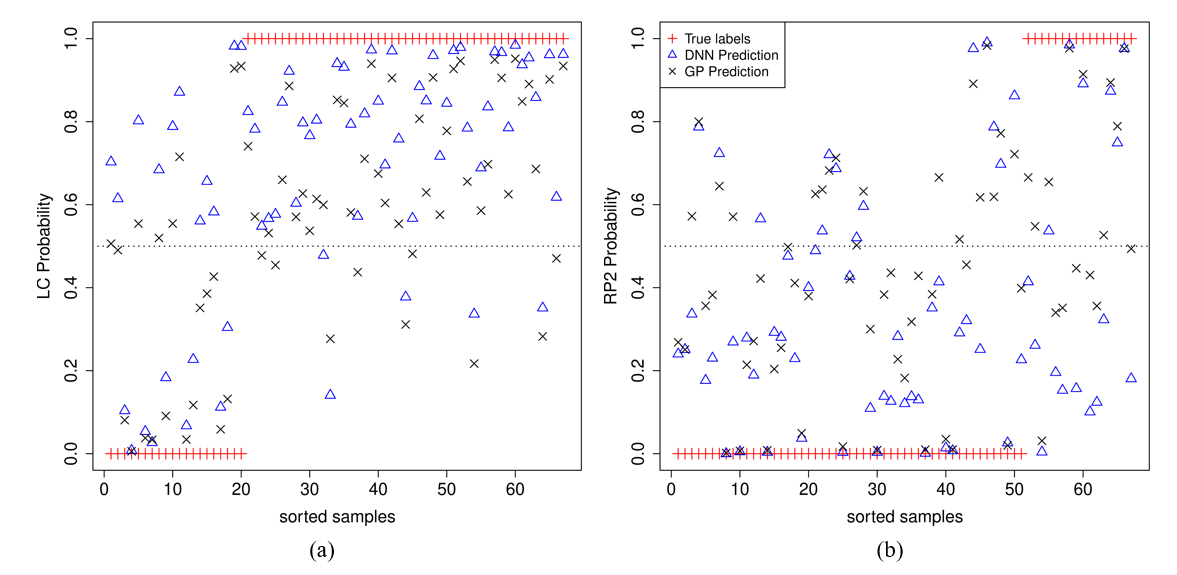}
    \caption{Comparisons between the predicted probabilities and the true binary LC/RP2 labels. The triangular represents the DNN predictions. The cross represents the prediction after GP calibration. The plus shows the true LC/RP2 labels. The dashed line depicts the $0.5$ probability threshold. (a) comparisons of the predicted LC probabilities. $20$ samples have LC=0 while $47$ samples have LC=1. (b) comparisons of the predicted RP2 probabilities. $51$ samples have RP2=0 while $16$ samples have RP2=1.} 
    \label{f:details}
\end{figure}

Next, the predicted week 6 patient variables are used to train the evaluation function. We assign a non-informative prior to the precision parameters $\lambda_{h_1}$ and $\lambda_{h_2}$ as
\begin{equation}
    p\left(\lambda_{h_j}\right)\propto \lambda_{h_j}^{-1}, j=1,2,
\end{equation}
maximize the likelihood in Equation~(\ref{eq:likelihood_evaluation}). through a grid search, and obtain the two precision parameters' values as $\lambda_{h_1}=\lambda_{h_2}=4$. Then the other GP parameters are estimated with the optimal precision parameters. With the parameters estimated, the LC and RP2 probabilities can be predicted. Figure~\ref{f:details} compares the detailed predicted probabilities on sorted samples. The GP model provides probabilities that are closer to the true binary labels. The cross-entropy \citep{goodfellow2016deep} is calculated for quantitative comparison. For LC, the GP model reduces the cross-entropy from $35.32$ to $33.58$. For RP2, the GP model reduces the cross-entropy from $52.19$ to $44.76$. More importantly, the GP model provides the confidence interval estimator of the LC and RP2 probabilities, which can be used to quantify the uncertainty of the treatment outcome based on physicians' prescriptions and AI recommendations.

\begin{figure}
    \centering
    \includegraphics[height=5.2in]{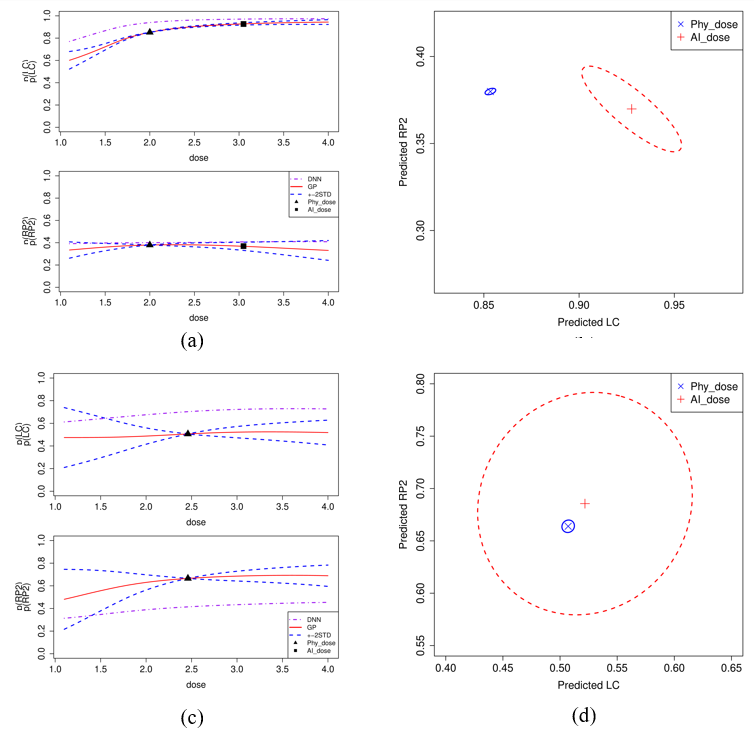}
    \caption{Uncertainty quantification of the treatment outcome for two patients. (a) Uncertainty quantification of the treatment outcome of different doses for the first patient. The solid, dashed, and dotted-dashed lines represent the GP prediction, $\pm2$ standard deviation confidence interval of GP prediction, and DNN prediction, respectively. The triangular depicts the physician's prescription. The square depicts the AI recommendation. (b) Predicted LC and RP2 probabilities for the first patient based on physician's prescription and AI recommendation. The plus represents the mean predictions of LC and RP2 based on the AI recommendation. The cross shows the mean predictions of LC and RP2 based on Physician's prescription. The two ellipses are the $\pm2$ confidence region of the LC/RP2 predictions. (c) Uncertainty quantification of the treatment outcome of different doses for the second patient. (d) Predicted LC and RP2 probabilities for the second patient based on physician's prescription and AI recommendation.}
    \label{f:uncertainty}
\end{figure}

As an illustration, the uncertainty quantification results of the treatment outcome for two patients are visualized in Figure~\ref{f:uncertainty}. GP enables our model to evaluate the uncertainty of the treatment outcome at different doses. At the observed data point, the uncertainty of the treatment outcome based on the AI recommendation is often greater than that from the physician's prescription. This is because GP has higher prediction accuracy around the point where the dose prescription and patient variables are known. For the first patient shown in Figure~\ref{f:uncertainty} (a) and (b), the AI algorithm suggests a higher dose prescription, which may result in better LC and similar RP2. To determine whether the AI recommendation or physician's prescription is better, we compute the p-value following the procedure in subsection~\ref{ss:integrate}. The resultant p-value of $0$ suggests to follow the AI recommendation for the future patients with $s_{T}$. For the second patient shown in Figure~\ref{f:uncertainty} (c) and (d), since the treatment outcome from AI recommendation has a large uncertainty, we would recommend to use physician's prescription.

\begin{figure}
    \centering
    \includegraphics[height=5in]{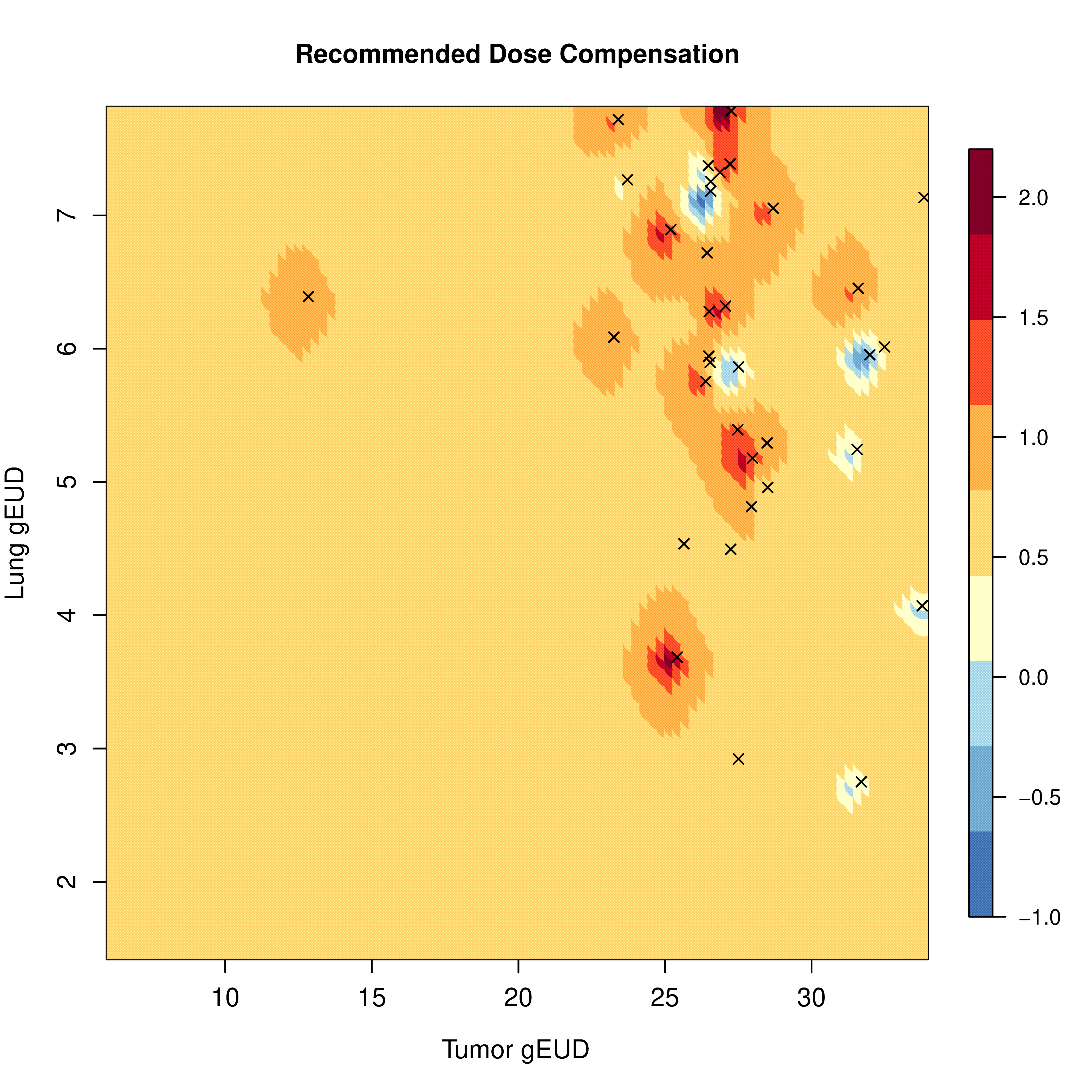}
    \caption{Visualization of the dose compensation for physicians' prescriptions. The regions with warm colors depict the combination of the patient variables that require higher doses comparing to the current physicians' prescriptions. The two patient variables have been transformed to their original scales for easier interpretation. The regions with cold colors implies that the physicians' should reduce their dose prescription for the patients with the corresponding patient variables. The black crosses illustrate the training samples.}
    \label{f:visualization}
\end{figure}

We conduct the analysis on all the $67$ patients. The integrative system suggests that the AI recommendation provides better treatment outcomes for $35$ patients based on the p-values. We further use these samples to predict the dose compensation, which may help physicians make better dose prescriptions. Since the physicians mainly determine the dose prescription based on the two patient variables ``Tumor gEUD'' and ``Lung gEUD'', we construct a Gaussian process to predict the dose compensation $a^A-a^P$ based on these two patient variables, and visualize the predictions in a 2-dimensional plot with respect to the two patient variables in Figure~\ref{f:visualization}. The warm background color implies that the physicians' prescriptions are overall conservative comparing to the recommended doses. It is also worth noting that the AI algorithm suggests to deliver higher doses for small Tumor\_gEUD, and lower doses for higher Tumor\_gEUD.
\section{Acknowledgement:}
The research is partly supported by NIH grant R01-CA233487.
\appendix
\section{Parameter estimation and Gaussian process prediction for the transition function}\label{aptransition}
Denote the observed patient variables at stage $t$ by $S_t$, and its $j$-th dimension by ${S_t}^{\left(j\right)}$. Denote the prescriptions at stage $t$ by $A_t$. The model parameters,  $\beta_\delta=\{\beta_{\delta,k,j},k=1,...,q+1,j=1,...,q\}$ and $\lambda_\delta=\{\lambda_\delta^{\left(j\right)},j=1,...,q\}$, can be trained by maximizing the log-likelihood as:
\begin{equation}
\left({\hat{\beta}}_\delta^{\left(j\right)},{\hat{\lambda}}_\delta^{\left(j\right)}\right)=arg\,max \log\,p\left(\beta,\lambda\middle| S_2^{\left(j\right)},\ldots,S_{T}^{\left(j\right)};S_1,\ldots,S_{T-1},A_1,\ldots,A_{T-1}\right),
\label{eq:gauss_est}
\end{equation}
with
\begin{eqnarray}
&\quad&\log\,p\left(\beta,\lambda\middle| S_2^{\left(j\right)},\ldots,S_{T}^{\left(j\right)};S_1,\ldots,S_{T-1},A_1,\ldots,A_{T-1}\right)\nonumber\\
&=&-\frac{1}{2}\sum_{t=1}^{T-1}\left(S_{t+1}^{\left(j\right)}-\eta_s^{\left(j\right)}\left(S_t^{\left(j\right)}\right)\right)^TK_\eta^{\left(j\right)^{-1}}((S_t,A_t),(S_t,A_t))\nonumber\\
&\times &\left(S_{t+1}^{\left(j\right)}-\eta_s^{\left(j\right)}\left(S_t^{\left(j\right)}\right)\right)-\frac{1}{2}\left|K_\delta^{\left(j\right)}((S_t,A_t),(S_t,A_t))\right|-\frac{n}{2}log2\pi,
\end{eqnarray}
where $K^{\left(j\right)}\left(\cdot,\cdot\right)$ is the Gram matrix of the covariance function $c_\delta^{\left(j\right)}\left(\cdot,\cdot\right)$. 

With the parameters trained, we can obtain the prediction of the $j-$th dimension of $s_{T+1}$, $s_{T+1}^{\left(j\right)}$. The prediction is expressed as a normally distributed random variable, whose distribution parameters are described as follows:
\begin{eqnarray}
s_{T+1}^{\left(j\right)}|s_{T},a_{T}&\sim&\mathcal{N}\left({\hat{\mu}}_T^{\left(j\right)}(s_T,a_T),{\widehat{\Sigma}}_T^{\left(j\right)}(s_T,a_T)\right),\nonumber\\
{\hat{\mu}}_{T+1}^{\left(j\right)}(s_T,a_T)&=&\eta_s^{\left(j\right)}\left(s_{T}\right)+{K^{\left(j\right)}}^T\left(s_{T},\cup_{t=1}^{T-1} S_{t}\right)K_\delta^{\left(j\right)^{-1}}\left(\cup_{t=1}^{T-1} S_{t},\cup_{t=1}^{T-1} S_{t}\right)\nonumber\\
&\times&\left(\cup_{t=1}^{T-1} S_{t+1}^{\left(j\right)}-\eta_s^{\left(j\right)}\left(\cup_{t=1}^{T-1} S_{t}\right)\right),\nonumber\\
{\widehat{\Sigma}}_{T+1}^{\left(j\right)}(s_T,a_T)&=&\frac{1}{\lambda_s^{\left(j\right)}}\{K_\delta^{\left(j\right)}\left(s_{T},s_{T}\right)-{K_\delta^{\left(j\right)}}^T\left(s_{T},\cup_{t=1}^{T-1} S_{t}\right)\nonumber\\
&\times& K_\delta^{\left(j\right)^{-1}}\left(\cup_{t=1}^{T-1} S_{t},\cup_{t=1}^{T-1} S_{t}\right)K_\delta^{\left(j\right)}\left(\cup_{t=1}^{T-1} S_{t},s_{T}\right)\}.
\label{eq:predictive_transition}
\end{eqnarray}
\section{Parameter estimation and Gaussian process prediction for the transition function}\label{apevaluation}
To predict the binary labels $Y_1$ and $Y_2$ without the observations of $h_1$ and $h_2$, we would like to marginalize the underlying logit of $LC$ and $RP2$ out, and hence calculate the densities of $h_1$ and $h_2$ via
\begin{equation}
    p\left(h_j\middle| \widehat{S}_{T+1},Y_j,s_{T+1}\right)=\int{p\left(h_j\middle| H_j,\widehat{S}_{T+1},Y_j,s_{T+1}\right)p\left(H_j\middle| \widehat{S}_{T+1},Y_j\right)dh_j},j=1,2,
    \label{eq:int_evaluate}
\end{equation}
where $\widehat{S}_{T+1}=\left\{\hat\mu_{T+1}^{(1)}(S_T),...,\hat\mu_{T+1}^{(q)}(S_T)\right\}$. The integration cannot be computed because the term $p\left(H_j\middle| \widehat{S}_{T+1},Y_j\right)$ is intractable. Therefore, Laplace approximation \citep{tierney1986accurate} is implemented to approximate $p\left(H_j\middle| \widehat{S}_{T+1},Y_j\right)$ with the normal density $\mathcal{N}\left(H_j\middle|{\widehat{H}}_j,\Lambda_j\right)$. The distribution parameters are calculated via maximizing the likelihood function $\Psi_j$ as
\begin{eqnarray}
    {\widehat{H}}_j&=&arg\,max_{h_j}\Psi_j\left(h_j\right),\nonumber\\
    \Lambda_j&=&\lambda_{h_j}{K_{h_j}}^{-1}\left(\widehat{S}_{T+1},\widehat{S}_{T+1}\right)+W_j,\nonumber\\
    W&=&\nabla\nabla\Psi_j\left(h_j\right),j=1,2.
\end{eqnarray}
The likelihood function $\Psi_j$ is defined as
\begin{eqnarray}
    \Psi_j\left(h_j\right)&=&\sum_{i=1}^{n}\left(y_{i,j}log\frac{\exp{h_{i,j}}}{1+\exp{h_{i,j}}}+\left(1-y_{i,j}\right)log\frac{1}{1+\exp{h_{i,j}}}\right)\nonumber\\
&-&\frac{\lambda_{h_j}}{2}h_j^TK_{h_j}^{-1}h_j+C.
\label{eq:likelihood_evaluation}
\end{eqnarray}
Here $\sigma\left(h\right)=exp\left(h\right)/\left(1+exp\left(h\right)\right)$, $K_{h_j}$ is the Gram matrices of the covariance function $c_{h_j}\left(\cdot,\cdot\right)$, $h_{i,j}$ is the $i$-th sample of $h_j$, and $C$ is some constant. Similar to the training process in Subsection~\ref{ss:transition}, the model parameters,  $\beta_h=\{\beta_{h,k,j},k=1,...,q+1,j=1,2\}$ and  $\lambda_h=\{\lambda_{h_1},\lambda_{h_2}\}$ can be estimated by maximizing the approximated probabilities $p\left(h_1\middle| \widehat{S}_{T+1},Y_1,s_{T+1}\right)$ and $p\left(h_2\middle| \widehat{S}_{T+1},Y_2,s_{T+1}\right)$ by substituting the $H_j$ into Equation~(\ref{eq:int_evaluate}) with ${\widehat{H}}_j$.
\section{Combination of the predictions of the transition and evaluation functions}\label{apcombination}
The confidence interval can be generated in two steps. First, we propagate the uncertainty of the estimated $s_T$ to that of the estimated $Prob[y_1]$ and $Prob[y_2]$, resulting in an error-in-variable GP model \citep{cressie2003spatial} for the logit functions of $h_1$ and $h_2$:
\begin{equation}
    h_j|s_{T},a\sim\mathcal{GP}\left({\hat{\mu}}_j,{\widehat{\Sigma}}_j\right),j=1,2,
\end{equation}
where
\begin{eqnarray}
{\hat{\mu}}_j&=&{\widetilde{K}}_{\ h_j}^T\left({\hat{\mu}}_{T+1},{\widehat{S}}_{T+1}\right){\widetilde{K}}_{h_j}^{-1}\left({\widehat{S}}_{T+1},{\widehat{S}}_{T+1}\right){\widehat{H}}_j,\nonumber\\
{\widehat{\Sigma}}_j&=&\frac{1}{\lambda_{h_j}}{\widetilde{K}}_{h_j}^T\left({\hat{\mu}}_{T+1},{\hat{\mu}}_{T+1}\right)\nonumber\\
&-&\frac{1}{\lambda_{h_j}}{\widetilde{K}}_{h_j}^T\left({\hat{\mu}}_{T+1},{\widehat{S}}_{T+1}\right){\widetilde{K}}_{h_j}^{-1}\left({\widehat{S}}_{T+1},{\widehat{S}}_{T+1}\right){\widetilde{K}}_{ h_j}\left({\hat{\mu}}_{T+1},{\widehat{S}}_{T+1}\right),
\label{eq:predictive_evaluation}
\end{eqnarray}
with ${\hat{\mu}}_{T+1}$ denoting the predicted mean of $s_{T+1}$ in Equation~(\ref{eq:predictive_transition}). ${\widetilde{K}}_{h_j}$ is the Gram matrix of the covariance function:
\begin{equation}
{\tilde{c}}_{h_j}\left(s,s\prime\right)=\prod_{k=1}^{q}\frac{1}{1+4\beta_{k,j}\sigma_{T+1,k}^2}exp{\left(-\frac{1}{1/\beta_{k,j}+4\sigma_{T+1,k}^2}\left(s^{\left(k\right)}-s^{\left(k\right)^\prime}\right)^2\right)},j=1,2,
\end{equation}
where $\{\sigma_{T+1,k}^2,k=1,...,q\}$ are the realization of ${\widehat{\Sigma}}_{T+1}^{\left(j\right)}$ in Equation~(\ref{eq:se_transition}) with input $s_{T}$. Applying the delta-method to Equation~(\ref{eq:predictive_evaluation}) yields the predictive distribution of the probabilities of LC and RP2, which are:
\begin{eqnarray}
Prob\left[y_j=1|s_{T},a\right]&\sim&\mathcal{N}\left(\frac{exp{\left(\hat{\mu}_j\right)}}{1+exp{\left(\hat{\mu}_j\right)}},D_j{\widehat{\Sigma}}_jD_j\right),\nonumber\\
D_j&=&diag\left\{\frac{exp{\left(\hat{\mu}_j\right)}}{\left(1+{exp{\left(\hat{\mu}_j\right)}}\right)^2}\right\}, j=1,2. 
\label{eq:sim}
\end{eqnarray}
\section{Detailed descriptions of the 12 selected patient variables}\label{appB}
\begin{itemize}
    \item \textbf{il4 (interleukin 4)}: Th2 cytokines: (i) Regulates antibody production, hematopoiesis, and inflammation. (ii) Promotes the differentiation of naïve helper T cells into Th2 cells. (iii) Decreases the production of Th1 cells \citep{ramirez2013perioperative, schaue2012cytokines, warltier2002systemic}. Here CD4 T helper cells are lymphocytes that strongly modulate the response of the immune system against cancer cells proliferation and tumor growth. They are classified into Th1 (antitumor) and Th2 (protumor) cells. Th1 and Th2 cells releases cytokines. Imbalance between Th1 and Th2 helps cancer cell.
    \item \textbf{il10 (interleukin 10)}: Th2 cytokines: (i) Inhibits synthesis of Th1 cytokines such as IFN-$\gamma$ (interferon gamma) and IL2. (ii) Inhibits antigen-presenting cells \citep{ramirez2013perioperative, schaue2012cytokines, warltier2002systemic}. \textbf{IFN-$\gamma$:} Th1 cytokines: (i) Enhances the microbicidal function of macrophages. (ii) Promotes the differentiation of naïve helper T cells into Th1 cells. (iii) Activates polymorphonuclear leukocytes, cytotoxic T cells, and NK cells. \textbf{IL2:} Th2 cytokines: (i) Promotes clonal expansion and development of T and B-lymphocytes. (ii) Induces expression of adhesion molecules. (iii) Enhances the function of NK cells.
    \item \textbf{il15 (interleukin 15)}: Th2 cytokines: (i) Induces activation and cytotoxicity of NK (natural killer) cells. (ii) Activates macrophages. (iii) Promotes proliferation and survival of T and B- lymphocytes and NK cells \citep{ramirez2013perioperative, schaue2012cytokines, warltier2002systemic}.
    \item \textbf{ip10 (interfeon gamma-induced protein 10):} IP-10 is secreted in response to IFN-$\gamma$ by various cells including monocytes, endothelial and fibroblasts. (i) Acts as chemoattractions for monocytes/macrophages, T cells, NK cells, and dendritic cells. (ii) Promotes T cell adhesion to endothelial cells. (iii) Antitumor activity (iv) Inhibition of bone marrow colony formation (v) Angiogenesis \citep{luster1985gamma, dufour2002ifn, angiolillo1995human}.
    \item \textbf{mtv:} Metabolic tumor volume from PET imaging.
    \item \textbf{GLSZM\_LZLGE:} Radiomics features: the large zone low gray-level emphasis (LZLGE) feature of a gray-level size zone matrix defined as $\sum_{i=1^{N_g}}\sum_{j=1}^{L_z}\frac{j^2p(i,j)}{i^2}$ \citep{carrier2018radiomics}.
    \item \textbf{GLSZM\_ZSV:} Radiomics features: the zone-size variance (ZSV) feature of a gray-level size zone matrix defined as $\frac{1}{N_gL_z}\sum_{i=1^{N_g}}\sum_{j=1}^{L_Z}(jp(i,j)-\mu_j)^2$ \citep{carrier2018radiomics}.
    \item \textbf{Lung gEUD ($\alpha/\beta=4Gy,10Gy\,resp$):} Generalized equivalent uniform dose of lung converted from EQD2 dose distributions: $EQD=N_{frac}\times d\times \left(\frac{d+\alpha/\beta}{2+\alpha/\beta}\right)$ \citep{bentzen2000morbidity, sovik2007parameter}.
    \item \textbf{Tumor gEUD ($\alpha/\beta=4Gy,10Gy\,resp$):} Generalized equivalent uniform dose of tumor converted from EQD2 dose distributions: $EQD=N_{frac}\times d\times \left(\frac{d+\alpha/\beta}{2+\alpha/\beta}\right)$ \citep{bentzen2000morbidity, sovik2007parameter}.
    \item \textbf{Rs2234671:} A SNP in the gene cxcr1 also known as Interleukin 8 receptor, alpha (IL8RA) relation to radiation induced toxicity in non-small cell lung cancer \citep{hildebrandt2010genetic}.
    \item \textbf{Rs238406:} A SNP in the gene ercc2 known to repair DNA excision and related to risk of lung cancer \citep{chang2008nucleotide}. \textbf{SNP:} Single nucleotide polymorphism is a substitution of single nucleotide that occurs at a specific position in the genome.
    \item \textbf{Rs1047768:} A SNP in the gene ercc5 also known to repair DNA excision and related to lung cancer susceptibility \citep{kiyohara2007genetic}.
\end{itemize}
\bibliographystyle{elsarticle-harv} 
\bibliography{mybib}





\end{document}